\documentclass[wcp,table,xcdraw]{jmlr}
\usepackage{times}
\usepackage{graphicx}
\usepackage{amsmath,amssymb} 
\usepackage{color}
\usepackage[]{algorithm2e}
\usepackage{newfloat}
\usepackage{listings}
\usepackage{algpseudocode}
\usepackage{xspace}
\usepackage{mathtools}
\usepackage{booktabs}
\usepackage{hhline}
\usepackage{wrapfig}
\usepackage{soul}
\newcommand{\name}{CAML\xspace}

\makeatletter
\let\Ginclude@graphics\@org@Ginclude@graphics 
\makeatother

\jmlrvolume{}
\jmlryear{}
\jmlrworkshop{Preprint}

\author{\Name{Rakshith Subramanyam} \Email{rsubra17@asu.edu}\\
   \addr Arizona State University
   \AND
   \Name{Mark Heimann} \Email{heimann2@llnl.gov}\\
   \addr Lawrence Livermore National Laboratory
   \AND
   \Name{Jayram Thathachar} \Email{Thathachar1@llnl.gov}\\
   \addr Lawrence Livermore National Laboratory
   \AND
   \Name{Rushil Anirudh} \Email{anirudh1@llnl.gov}\\
   \addr Lawrence Livermore National Laboratory
   \AND
   \Name{Jayaraman J. Thiagarajan} \Email{jjayaram@llnl.gov}\\
   \addr Lawrence Livermore National Laboratory
   }

\title[Contrastive Knowledge-Augmented Meta-Learning for Few-Shot Classification]{Contrastive Knowledge-Augmented Meta-Learning for Few-Shot Classification }

\begin{document}

\maketitle
\thispagestyle{empty}

\begin{abstract}
Model agnostic meta-learning algorithms aim to infer priors from several observed tasks that can then be used to adapt to a new task with few examples. Given the inherent diversity of tasks arising in existing benchmarks, recent methods use separate, learnable structure, such as hierarchies or graphs, for enabling task-specific adaptation of the prior. While these approaches have produced significantly better meta learners, our goal is to improve their performance when the heterogeneous task distribution contains challenging distribution shifts and semantic disparities. To this end, we introduce \name~(\ul{C}ontrastive Knowledge-\ul{A}ugmented \ul{M}eta \ul{L}earning), a novel approach for knowledge-enhanced few-shot learning that evolves a knowledge graph to effectively encode historical experience, and employs a contrastive distillation strategy to leverage the encoded knowledge for task-aware modulation of the base learner. Using standard benchmarks, we evaluate the performance of \name~in different few-shot learning scenarios. In addition to the standard few-shot task adaptation, we also consider the more challenging multi-domain task adaptation and few-shot dataset generalization settings in our empirical studies. Our results shows that \name~consistently outperforms best known approaches and achieves improved generalization.
\let\thefootnote\relax\footnote{This work was performed under the auspices of the U.S. Department of Energy by the Lawrence Livermore National Laboratory under Contract No. DE-AC52-07NA27344, Lawrence Livermore National Security, LLC.and was supported by the LLNL-LDRD Program under Project No. 21-ERD-012}
\end{abstract}
\section{Introduction}
Learning to solve new tasks using only few-shot examples is a long-standing challenge. Meta-learning forms an important class of few-shot learning algorithms that leverages transferable knowledge priors from previously learned tasks to learn new tasks quickly. For example, the widely adopted model-agnostic meta-learning (MAML) approaches~\citep{finn2017model,yoon2018bayesian,lee2018gradient,finn2018probabilistic,finn2017meta} attempt to learn a single \textit{meta} model (or base learner) on a set of observed tasks, which is assumed to be only a few gradient descent steps away from good task-specific models. Their success hinges on the assumption that the observed tasks are realizations from a common task distribution $p(\mathcal{T})$. Despite its mathematical tractability, the premise of using a single base learner can be insufficient when $p(\mathcal{T})$ is heterogeneous, \textit{i.e.}, the degree of similarity between tasks can be vastly different~\citep{vuorio2019multimodal}. This motivates the need for a meta-model to selectively utilize knowledge from its previous experience that is the most relevant for the target task. 

In this context, task-aware modulation~\citep{vuorio2019multimodal} (MuMo-MAML) is an important principle to improve the performance of MAML on heterogeneous tasks. Conceptually, MuMo-MAML infers latent representations to characterize realizations from a heterogeneous task distribution, and modulates the base learner initialization appropriately, based on latent characteristics, to improve the adaptation performance. In a quest to improve upon MuMo-MAML, recent methods have resorted to using a separate learnable structure for encapsulating relationships across the training episodes~\citep{vuorio2019multimodal,yao2019hierarchically,yao2020automated}. This meta structure can then be leveraged when adapting to a new task by selectively utilizing knowledge from prior experience. Though these approaches have been found to outperform vanilla MuMo-MAML in few-shot adaptation tasks, they have the risk of overfitting the \textit{task encoding network} to the observed distribution of training tasks. As we demonstrate in this paper, this behavior can be particularly limiting in the more challenging (yet practical) settings of multi-domain task adaptation~\citep{peng2019moment} and few-shot dataset generalization~\citep{triantafillou2021learning}.

\begin{figure*}[t]
    \centering
    \includegraphics[width=0.9\textwidth]{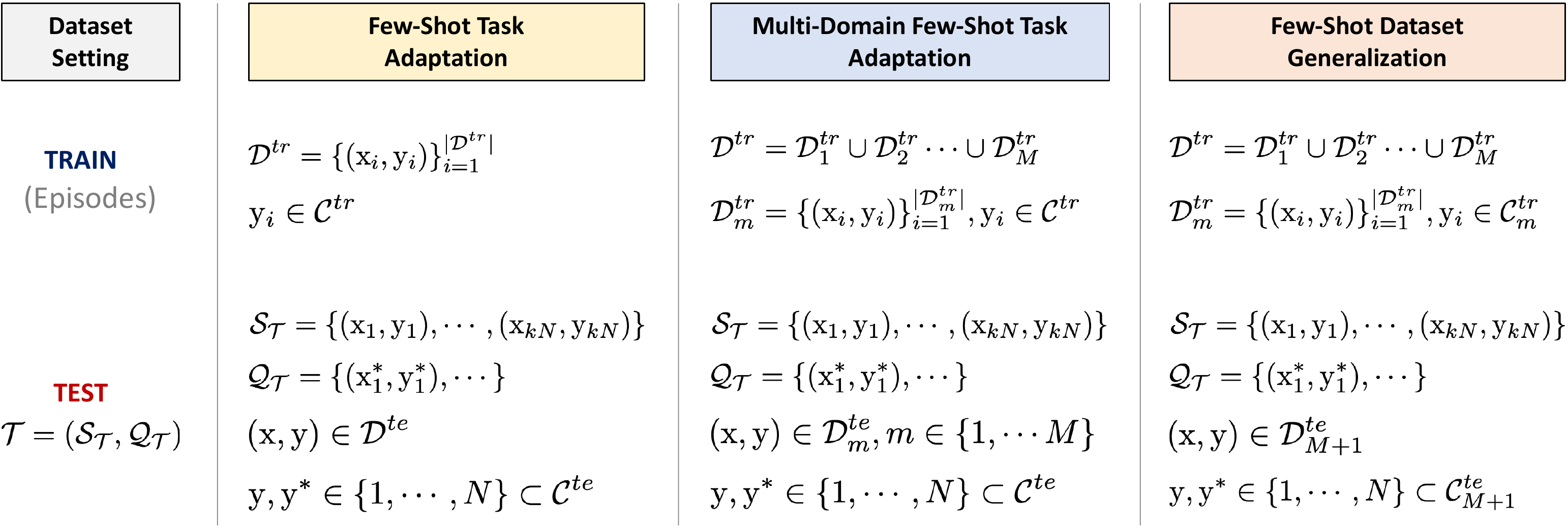}
    \caption{\textbf{Few-shot classification tasks}. Here, we formally define the different problem settings considered in this study. As we move from few-shot adaptation to few-shot dataset generalization, the problem becomes increasingly challenging and requires sophisticated task-aware modulation strategies to improve the performance of MAML.}
    \label{fig:problems}
\end{figure*}

\noindent \textbf{Proposed Work.} We introduce a new approach, Contrastive Knowledge-Augmented Meta Learning (\name), for task-aware modulation in MAML with the goal of advancing the generalization capabilities of meta-learners. At its core, CAML utilizes prototype graphs to represent a few-shot task and leverages an external meta knowledge structure to construct task embeddings, which are subsequently used to modulate the base learner parameters. CAML constructs the meta knowledge structure from the observed tasks similar to state-of-the-art methods~\citep{yao2020automated}, albeit with the following key differences: (i) CAML uses a contrastive knowledge distillation strategy to systematically infuse prior knowledge directly into the image embedding module, which leads to richer task representations and eliminates the need to perform knowledge extraction during inference time; (ii) Through the improved image embeddings, CAML enables a task encoding scheme without any learnable parameters (average pooling) to construct task representations; in contrast, state-of-the-art approaches rely on complex task encoding modules; (iii) CAML employs a moving average-based update strategy for the knowledge structure, which leads to improved encoding of the prior experience compared to existing approaches~\citep{yao2019hierarchically,yao2020automated}. 

We evaluate \name~and compare to other few-shot learning methods using three experimental settings (see Figure \ref{fig:problems}): (i) the standard few-shot adaptation setting, where the goal is to adapt to novel classes within the observed datasets; (ii) multi-domain adaptation, where the goal is to learn priors using data sampled from different domains and adapt to unseen test tasks sampled from any of the observed domains; and (iii) few-shot dataset generalization, where the test tasks are sampled from novel unseen datasets that are not observed during training. We make an interesting finding that, even without using the knowledge structure for inference, \name is able to produce highly effective task-specific initializations. While \name~is very competitive in the standard task adaptation setting, in the multi-domain setting, its benefits become even more apparent, and we observe larger performance gains ($2\% - 3\%$ for both $1-$shot and $5-$shot training). Finally, in the most challenging dataset generalization setting, we find that \name~significantly outperforms the state-of-the-art structure-aware learning method (ARML~\citep{yao2020automated}), with gains as high as $~5\%$ and $~10\%$ in $1$- and $5$-shot training respectively.
\section{Problem Setup}

In this section, we describe the problem settings considered in this study for studying the behavior of different task-aware meta learning approaches. Figure~\ref{fig:problems} provides an overview of the formulations considered. Broadly, in few-shot classification, training tasks drawn from the distribution $p^{tr}(\mathcal{T})$ are used to learn how to adapt quickly to any of the tasks, and evaluated on previously unseen test tasks from $p^{te}(\mathcal{T})$. Common to all these formulations is that within each of the one or more datasets, the classes seen during training are completely disjoint from those seen during testing. However, the formulations vary in terms of distribution shifts and semantic disparities.

\vspace{0.1in}
\noindent \textbf{A. Few-shot Task Adaptation.} In this setup, let $\mathcal{D}^{tr} = \{(\mathrm{x}_i, \mathrm{y}_i)\}_{i=1}^{|\mathcal{D}^{tr}|}$ denote the training set comprising samples $\mathrm{x}_i$ and their labels $\mathrm{y}_i$, where $\mathrm{y}_i \in \mathcal{C}^{tr}$. In other words, all samples used for training belong to one of the classes from the set $\mathcal{C}^{tr}$. The goal is to learn an adaptable model using $\mathcal{D}^{tr}$ to support learning new classes with only few examples. For evaluation, we construct a series of few-shot tasks (or episodes) and measure the model's ability to adapt to novel classes from $\mathcal{C}^{te}$ (i.e., $\mathcal{C}^{tr} \cap \mathcal{C}^{te} = \emptyset$). More specifically, each $k-$shot $N-$way test episode is represented as the tuple $\mathcal{T} = (\mathcal{S}_{\mathcal{T}}, \mathcal{Q}_{\mathcal{T}})$, where the \textit{support} set contains $k$ examples from each of the $N$ classes (selected from $\mathcal{C}^{te}$), i.e., $\mathcal{S}_{\mathcal{T}} \coloneqq \{(\mathrm{x}_1, \mathrm{y}_1), \cdots, (\mathrm{x}_{kN}, \mathrm{y}_{kN})\}$, $\mathrm{y}_i \in \{1,\cdots, N\}$, and the \textit{query} set $\mathcal{Q}_{\mathcal{T}} \coloneqq \{(\mathrm{x}^*_1, \mathrm{y}^*_1), \cdots\}$ contains different (unlabeled) examples from the same set of $N$ classes.

\vspace{0.1in}
\noindent \textbf{B. Multi-Domain Few-shot Task Adaptation.} In many practical applications, the training examples $\mathrm{x}_i \in \mathcal{D}^{tr}$ can encompass a variety of distribution shifts (e.g., variations in the data acquisition process). Hence, in this setting, we represent the training set as a composition of datasets from $M$ different domains, i.e., $\mathcal{D}^{tr} = \mathcal{D}^{tr}_1 \cup \mathcal{D}^{tr}_2 \cdots \cup \mathcal{D}^{tr}_M$, wherein all samples (regardless of the domain) belong to a 
common family of classes $\mathcal{C}^{tr}$. The goal here is to learn to adapt to the training tasks arising from any of the $M$ domains. For evaluation, both the support and query sets for a test episode $\mathcal{T}$ are drawn from any domain $m \in \{1,\dots,M\}$, i.e., $(\mathrm{x}, \mathrm{y}) \in \mathcal{D}^{te}_m$ and the $N$ classes are picked from a disjoint set $\mathcal{C}^{te}$ similar to the previous case.

\begin{figure}[t]
    \centering
    \includegraphics[width=0.7\columnwidth]{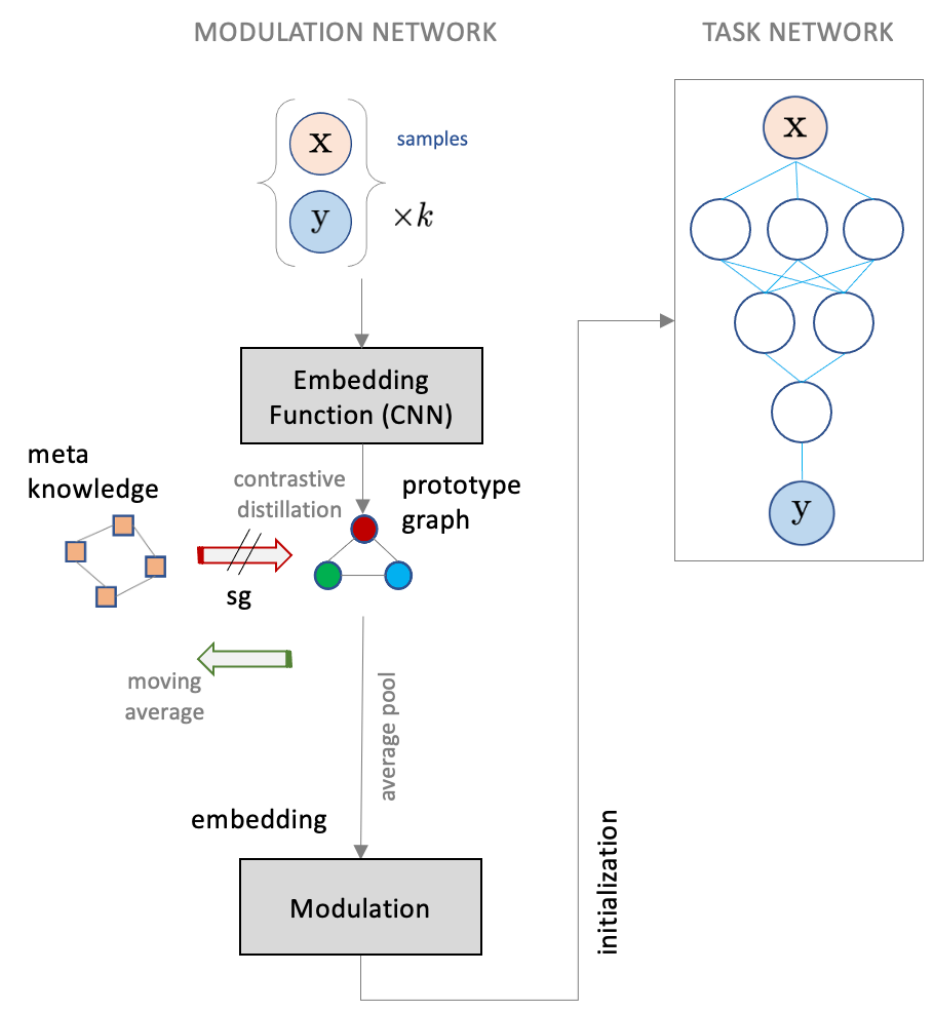}
    \caption{\textbf{Approach Overview}. An illustration of the proposed approach for task-aware meta learning.}
    \label{fig:overview}
\end{figure}

\vspace{0.1in}
\noindent \textbf{C. Few-shot Dataset Generalization.} In this challenging variant, the training set is defined as a union of $M$ different datasets $\mathcal{D}^{tr} = \mathcal{D}^{tr}_1 \cup \mathcal{D}^{tr}_2 \cdots \cup \mathcal{D}^{tr}_M$, and more importantly, it is assumed that each dataset contains examples from different sets of classes $\{\mathcal{C}^{tr}_m\}_{m=1}^M$. As a result, the goal here is to learn to adapt to completely different semantic concepts corresponding to each of the $M$ datasets. For evaluation, we construct test episodes using novel unseen classes from an entirely different dataset $\mathcal{D}_{M+1}^{te}$. Denoting the set of classes in the novel dataset as $\mathcal{C}_{M+1}^{te}$, we will study how effectively one can leverage the prior to generalize to unseen datasets.

\section{Task-Aware Meta Learning}

While the few-shot learning literature encompasses a wide variety of training algorithms, meta-learning is a popular choice~\citep{thrun2012learning} for both classification and reinforcement learning~\citep{nagabandi2018learning}. Existing few-shot meta-learning approaches can be broadly categorized into: 1) metric-based meta-learning frameworks~\citep{snell2017prototypical,koch2015siamese,vinyals2016matching} that learn a metric or distance function to compare different exemplars; 2) model-based approaches where meta-learning models learn to adjust the model parameters to adapt to new tasks~\citep{munkhdalai2017meta,santoro2016meta}; and 3) gradient-based model agnostic meta-learning models. In particular, our work builds upon model agnostic meta-learning (MAML)~\citep{finn2017model} formulated as follows:

\paragraph{\textbf{Model Agnostic Meta-Learning.}} Given a set of episodes, $\{\mathcal{T}^{tr}_1, \cdots, \mathcal{T}^{tr}_R\}$ comprised of support and query sets ($\mathcal{T}^{tr}_i = (\mathcal{S}_{\mathcal{T}^{tr}_i},\mathcal{Q}_{\mathcal{T}^{tr}_i})$), from the training set $\mathcal{D}^{tr}$, MAML considers the meta-learner as the initialization of a task network $f$, \textit{i.e.}, $\theta_0$, and optimizes for a well-generalized initialization $\theta_0^*$. Formally, 
\begin{align}
    \theta_0^* &= \arg \min_{\bar{\theta}} \sum_{i=1}^R \mathcal{L}(f_{\theta_i}; \mathcal{Q}_{\mathcal{T}^{tr}_i}) \\ 
    &= \arg \min_{\bar{\theta}} \sum_{i=1}^R \mathcal{L}(f_{\bar{\theta} - \alpha \nabla_{\theta} \mathcal{L}(\theta; \mathcal{S}_{\mathcal{T}^{tr}_i})\lvert_{\theta=\bar{\theta}}}; \mathcal{Q}_{\mathcal{T}^{tr}_i}),
    \label{eq:meta}
\end{align}where the task-specific initialization $\theta_i$ is obtained using a gradient step from the meta-initialization $\theta_0$. Note, the notation $\bar{\theta}$ refers to the variables used during the optimization of this bi-level objective function. Here, $\mathcal{L}(\theta; \mathcal{S}_{\mathcal{T}^{tr}_i})$ is implemented as the cross entropy loss $\sum_{(\mathrm{x}, \mathrm{y})\in \mathcal{S}_{\mathcal{T}^{tr}_i}} \log P(\mathrm{y} | \mathrm{x}, f_{\theta})$.

\vspace{0.1in}
\noindent \textbf{\textbf{Task-Aware Modulation.}} When the tasks used for meta-learning are sampled from a heterogeneous task distribution, inferring a common parameter initialization $\theta_0$ for all tasks can be fundamentally restrictive. Hence, task-aware modulation~\citep{vuorio2019multimodal} is a more effective formulation that aims at building a meta-learner which can generalize on heterogeneous task distributions through a set of latent parameters representing task-specific characteristics. For example, MuMo-MAML first uses a task encoder to encode the training episode for a given task into a task embedding vector $\mathrm{v}_i$. The task embedding is then used to obtain modulation vectors that are applied to the global initial parameters $\theta_0$ thereby producing task-aware initialization $\theta_{0i}$. Extending the MAML formulation in \eqref{eq:meta}, the task-aware modulation can be carried out using the support set in the training episode $\mathcal{S}_{\mathcal{T}^{tr}_i}$ and the updated initialization $\theta_{0i}$ is used to perform the meta-optimization.

While task-specific initialization can lead to improved generalization on heterogeneous tasks, its effectiveness relies on the ability of the task embeddings to encapsulate all relationships between the large number of observed tasks and thus performing similar parameter modulation for related tasks. Since it is challenging to learn such expressive embeddings, more recent approaches have resorted to storage and retrieval of task-relevant information from historical experience, in order to better balance generalization and customization (task-aware modulation)~\citep{yao2019hierarchically,yao2020automated}. For example, the recently proposed hierarchically structured meta learning (HSML) and automated relational meta-learning (ARML) approaches~\citep{yao2020automated} use an external meta knowledge structure to assist the task encoding process. By adopting these knowledge-enhanced representations coupled with a sophisticated RNN-based task encoder, these approaches are known to outperform MAML and MuMo-MAML in the standard, few-shot task adaptation setting.

\vspace{0.1in}
\noindent \textbf{\textbf{Our Approach.}} In this paper, our goal is to improve the performance of task-aware modulation strategies in the more challenging multi-domain task adaptation and dataset generalization settings. To this end, we develop \name (see Figure \ref{fig:overview}), a novel task-aware modulation approach that improves upon MuMo-MAML by: (i) evolving a meta knowledge structure $\mathcal{M}$, similar to state-of-the-art approaches; (ii) adopting a contrastive distillation strategy (with graph neural networks) to incorporate information from $\mathcal{M}$ to enrich the image embedding module; (iii) leveraging the improved image embeddings through the use of a simple task encoding scheme without any learnable parameters, unlike existing approaches that use complex task encoders; and (iv) utilizing a moving average-style update of $\mathcal{M}$ to effectively aggregate the historical experience. Interestingly, we find that, when adapting the meta-learned model to an unseen test task, \name~does not require extracting information from $\mathcal{M}$, unlike existing approaches~\citep{yao2019hierarchically,yao2020automated}.

\section{Algorithm}
In this section, we present our approach for task-aware modulation in MAML.

\subsection*{Step 1: Prototype Graph Generation}
Conventionally, feature extractors are used to embed data in low-dimensional latent spaces, where the different classes are easily separable. In task-aware modulation, our goal is to obtain such representations for different few-shot tasks, such that two tasks that are similar in the latent space can use the same task network initialization for effective adaptation. Following the notations introduced in the previous section, each $k-$shot $N-$way training episode $\mathcal{T}_i^{tr}$ is comprised of support and query sets $(\mathcal{S}_{\mathcal{T}_i^{tr}} \mathcal{Q}_{\mathcal{T}_i^{tr}})$, wherein there are $k$ samples in each of the $N$ classes randomly selected from $\mathcal{C}^{tr}$. Since effective task encoding is essential for suitably modulating the initialization of the task network (see Figure \ref{fig:overview}), \name~begins by constructing a prototype-based graph \citep{yao2020automated} with the image embeddings. 

Formally, given the support set $\mathcal{S}_{\mathcal{T}_i^{tr}} \coloneqq \{(\mathrm{x}_j, \mathrm{y}_j), \forall j \in [1, \cdots, kN]\}$ for a training episode, we compute embeddings for each image $\mathrm{x}_j$ in the task using an embedding function. 
While a variety of design choices can be adopted for this, we implement the embedding function using a ResNet-18 architecture.
Given that these embeddings are used to only perform the modulation (\textit{i.e.}, the task network does not utilize these embeddings), even a shallow CNN network can be used. Using the sample-level embeddings, we then compute the prototype vector for each class $n \in [1, \cdots, N]$ by taking the average of the embeddings:
\begin{equation} \label{eq:proto_nodes}
    \mathrm{v}^n_i = \frac{1}{k} \sum_{\substack{(\mathrm{x}_j, y_j) \in \mathcal{S}_{\mathcal{T}_i^{tr}} \\ y_j = n}} \mathcal{B}(\mathrm{x}_j),
\end{equation} where $\mathcal{B}$ denotes the feature extractor that projects an image $\mathrm{x}_j$ into $\mathbb{R}^d$. Given the sensitivity of few-shot learning methods to the limited number of examples, operating on the prototype representations reduces the effect of atypical samples. While the prototype graph representation is used to both optimize the knowledge-aware task encoding and to update the meta knowledge graph, for the final task encoding, we adopt an unweighted average of node embeddings from the prototype graph.
\begin{equation}\label{eq:task_rep_vec}
    \mathrm{z}_i = \mathrm{\Psi}(\mathcal{S}_{\mathcal{T}_i^{tr}}) =  \frac{1}{N} \sum_n \mathrm{v}_i^n
\end{equation}

Node embeddings with high class separability and infused prior knowledge enables the use of such simple feature aggregation strategies in contrast to  state-of-the-art approaches such as ARML~\citep{yao2020automated} and HSML~\citep{yao2019hierarchically}, which require sophisticated aggregation strategies (e.g., RNN autoencoders) for effective task-aware modulation. 



\subsection*{Step 2: Knowledge-Enhanced Task Encoding}
\label{sec:kd}
The desideratum of an ideal embedding function in task-aware modulation is to produce expressive task representations that capture the complexity of a given task. While this embedding function can be directly optimized with the goal of improving the adaptation performance with validation tasks, as done by MuMo-MAML~\citep{vuorio2019multimodal}, we propose to leverage historical experience (that evolves as we observe more training episodes) to enrich the task encoding. To this end, we adopt a meta knowledge graph structure to encode the historical experience and propose a novel contrastive distillation strategy to produce knowledge-enhanced task encodings. Note that existing structured meta-learning algorithms update the meta knowledge structure (a hierarchy or a graph) directly using gradients from the meta update step, which limits its ability to learn class-specific characteristics from the task at hand. Instead, in our approach, we do not allow gradients to directly alter the meta knowledge graph (see Figure \ref{fig:overview}), but employ an exponential moving average update based on the current task representations. As we will show in our experiments, this improved encoding of the historical experience leads to significantly improved few-shot generalization, particularly under challenging shifts.

Formally, denoting the knowledge graph for encoding experience as $\mathcal{M}$ with randomly initialized node features $\mathcal{H}_{\mathcal{M}} = \{\mathrm{h}_j\}, j = 1, \cdots, M$ and edges $\mathcal{E}_{\mathcal{M}}$, we will now describe our knowledge-aware task encoding process. Using the prototype graph from the previous step, $\mathcal{G}_i = (\mathcal{V}_i, \mathcal{E}_i)$, we perform a contrastive distillation from the meta knowledge graph $\mathcal{M}$. While the node features of the prototype graph are obtained using the embedding function $\mathcal{B}$ (as described in Step 1), those in the meta knowledge graph persist across episodes and are computed using a combination of neural message passing and an exponential moving average, to be described in Step 4. The edges in both these graphs are parameterized as a function of the absolute difference of the corresponding node features. For example, for any two nodes with features $\mathrm{a}$ and $\mathrm{b}$, $Edge(\mathrm{a},\mathrm{b}) = \sigma(\mathrm{U}^T|\mathrm{a} - \mathrm{b}|)$, where $\mathrm{U} \in \mathbb{R}^{d\times 1}$ denotes the learnable parameters and $\sigma$ is the sigmoid function. Note that for all pairs of nodes within a graph we learn a common weight matrix $U$.

In order to extract information for an episode $\mathcal{T}_i$ from historical experience, we construct a super-graph comprising nodes from both $\mathcal{G}_i$ and $\mathcal{M}$. The cross-edges between the two graphs are computed as the softmax of the set of negative Euclidean distances between all the pairs (no learnable parameters). For a pair $\mathrm{v}_i^n$ from $\mathcal{V}_i$ and $\mathrm{h}_j$ from $\mathcal{M}$,
\begin{equation}\label{eq:cross-edge}
Edge(\mathrm{v}_i^n, \mathrm{h}_j) = \frac{\exp(-\|\mathrm{v}_i^n - \mathrm{h}_j\|_2^2/\gamma)}{\sum_{\bar{n}, \bar{j}}\exp(-\|\mathrm{v}_i^{\bar{n}} - \mathrm{h}_{\bar{j}}\|_2/\gamma)}.
\end{equation}

In our implementation, we set $\gamma = 8$. In order to effectively balance between knowledge-enhanced representations and the native representations from the embedding function, we propose a contrastive learning strategy inspired by several existing self-supervised learning approaches such as SimCLR and InfoNCE~\citep{simclr,infonce}. 
Here, we consider the positive pair to be the task encodings from the original prototype representations and the knowledge-enhanced prototype representations obtained via neural message passing on the super-graph. On the other hand, the negatives correspond to the node pairs from $\mathcal{G}_i$, which indicate the level of class separability in the prototype graph. This objective $ \mathcal{L}_{CKD}(\mathcal{T}_i)$ can be expressed as:
\begin{equation}\label{eq:CKD}
- \mathbb{E}\bigg[\log \frac{\exp(\texttt{sim}(\mathrm{z}_i,\hat{\mathrm{z}}_i))}{\exp(\texttt{sim}(\mathrm{z}_i,\hat{\mathrm{z}}_i)) + \sum \exp(\texttt{sim}(\mathrm{v}_i^m, \mathrm{v}_i^n))} \bigg].
\end{equation}Here, $\hat{\mathrm{z}}_i = \mathrm{\Psi}[NMP(\mathcal{G}_i,\mathcal{M})]$ indicates the knowledge-enhanced task representations obtained by first performing neural message passing (NMP) on the super-graph (similar to~\citep{yao2020automated}) and then subsequently averaging the updated prototype node representations $\hat{\mathrm{v}}_i^n$. Note that, when performing neural message passing to obtain knowledge-
enhanced task representations, we do not allow the node features in the meta knowledge graph $\mathrm{h}_j$ to be changed, and only the prototype representations are updated. Furthermore, the similarity function $\texttt{sim}$ is implemented using the cosine similarity. In effect, this attempts to modify the embedding function such that the task encoding is consistent with the historical experience while also maximizing the inter-class separability, thereby producing highly effective task representations. 

\subsection*{Step 3: Task-Specific Modulation}
The next step of our approach is to utilize the knowledge-enhanced task representations for inferring a task-specific meta initialization. The inductive biases from the historical knowledge can enable rapid adaptation of the global meta-model to a novel task. To this end, the task representation $\mathrm{z}_i$ is used to implement the following modulation function on the global task network initialization $\theta_{0}$:
\begin{equation} \label{eq:modulation}
    \theta_{0i} = \mathrm{\Gamma}(\theta_0) = \sigma(\mathbf{W}_g\mathrm{z}_i + \mathrm{b}_g) \circ \theta_0,
\end{equation}where $\mathbf{W}_g,\mathrm{b}_g$ are learnable parameters. Using a gradient-through-gradient optimization, one can then refine the task-specific initialization $\theta_{0i}$. We incorporate our distillation objective from step 2 into this task-aware modulation process from MuMo-MAML:
\begin{equation}
\label{eq:proposed}
    \min_{\bar{\theta}, \Omega} \sum_{i=1}^R \mathcal{L}(f_{\mathrm{\Gamma}(\bar{\theta}) - \alpha \nabla_{\theta} \mathcal{L}(\theta; \mathcal{S}_{\mathcal{T}^{tr}_i})\lvert_{\theta=\mathrm{\Gamma}(\bar{\theta})}}; \mathcal{Q}_{\mathcal{T}^{tr}_i}) + 
    \\ \lambda \mathcal{L}_{CKD}(\mathcal{T}_i).
\end{equation} Here, $\Omega$ corresponds to the parameters of feature extractor $\mathcal{B}$, NMP network and modulation function $\Gamma$. The hyper-parameter $\lambda$ controls the influence of the contrastive distillation term in the overall objective.

\subsection*{Step 4: Meta Knowledge Graph Update}
The final step in \name~is to evolve the meta knowledge graph $\mathcal{M}$ with the current task information. As described earlier, by not allowing gradients from the meta update step to alter node features $\mathcal{H}_{\mathcal{M}}$, we are able to better control the historical experience encoded in $\mathcal{M}$. More specifically, using the dataset $\mathcal{T}^{tr}_i$ for each $i$ in parallel, we update the node features $\mathrm{h}_j \in \mathcal{H}_{\mathcal{M}}$ using neural message passing on the super-graph to obtain $\hat{\mathrm{h}}^i_j$. In contrast to the distillation loss computation, during this NMP, we do not allow the prototype node features to be changed and update only the node features of $\mathcal{M}$. 
Note that, for both the prototype and the knowledge graphs, we use the edges inferred after the meta update in Step~3. Let $\hat{\mathrm{h}}_j$ denote the average of all the $\hat{\mathrm{h}}^i_j$. Finally, we employ an exponential moving average update of the node features of the meta knowledge graph via $\mathrm{h}_j = \alpha \hat{\mathrm{h}}_j + (1 - \alpha) \mathrm{h}_j$, where the hyper-parameter $\alpha$ controls the amount of history retained from previous episodes.

    
\begin{algorithm2e}[t]
    \SetAlgoLined
    \DontPrintSemicolon
    \caption{~Training of \name}\label{algo1}
    
	\textbf{Input}: $p^{tr}(\mathcal{T})$ Distribution over training tasks, hyper-parameters $\alpha,\lambda$ \;
	
	\textbf{Learnable Parameters}: Embedding network $\mathcal{B}$, task network $f(\theta_0)$, modulation parameters $\mathrm{\Gamma}$, meta knowledge graph $\mathcal{M}$, NMP network\;
	
	\textbf{Initialization}: Randomly initialize parameters $\theta_{0}$, $\mathcal{B}$, $\mathcal{M}$, and NMP network
	 \;
    
    \While{not done}{
        Sample a batch of tasks $\mathcal{T}_i^{tr} \sim p^{tr}(\mathcal{T})$;
            
            //meta-train // \\
            \For {each $\mathcal{T}_i^{tr}$}{
                Sample $\mathcal{S}_{\mathcal{T}_i^{tr}}$ and$ \mathcal{Q}_{\mathcal{T}_i^{tr}}$ from $\mathcal{T}_i^{tr}$\;

                Randomly initialize learnable edges of $\mathcal{M}$ \;
                
                Compute the prototype vectors $\mathcal{V}_i$ following equation \eqref{eq:proto_nodes} \;
                
                 Construct prototype graph $\mathcal{G}_i$ with learnable edges \;
                
                Construct the task representation vector $\mathrm{z}_i$ following equation \eqref{eq:task_rep_vec}\;

                Compute the contrastive objective $\mathcal{L}_{CKD}(\mathcal{T}_i)$ following equation \eqref{eq:CKD} \;

                Perform task-aware modulation using $\mathrm{z}_i$ following equation \eqref{eq:modulation} \;
                
             Update 
            $\theta^{*}_0 = \theta_0  - \alpha \nabla_{\theta} \mathcal{L}(\theta; \mathcal{S}_{\mathcal{T}^{tr}_i})$ \;
            }
            //meta-update// \\
            Minimize the objective in \eqref{eq:proposed} and update $\theta_0$, $\mathcal{B}$, $\mathrm{\Gamma}$, edge weights of $\mathcal{M}$, and NMP network\;
            
            \For {each $\mathcal{T}_i^{tr}$}{
            Obtain $\hat{\mathcal{H}}^i_{\mathcal{M}}$ with $\mathcal{G}_i$ using the strategy in Step 4 of Section 4. 
            }
            Update $\mathcal{M}$ using $\hat{\mathcal{H}}_{\mathcal{M}}$ averaged over $\mathcal{T}_i^{tr}$
    }

\end{algorithm2e}

\section{Results and Findings}
\begin{table*}[t]
\centering
\caption{\textbf{Few-shot task adaptation}. Performance comparison of the proposed approach against state-of-the-art meta-learning methods. Following standard practice, we show results for $4$ different datasets from the Meta-Dataset benchmark. }
\renewcommand*{\arraystretch}{1.3}
\resizebox{0.9\textwidth}{!}{
\begin{tabular}{|c||c||c||c||c||c|}
  \hhline{|-||-||-||-||-||-|}
  \cellcolor[HTML]{C0C0C0}Method& 
  \cellcolor[HTML]{C0C0C0}Bird&
  \cellcolor[HTML]{C0C0C0}Texture&
  \cellcolor[HTML]{C0C0C0}Aircraft & 
  \cellcolor[HTML]{C0C0C0}Fungi & 
  \cellcolor[HTML]{C0C0C0}Average\\
\hhline{======}
\rowcolor[HTML]{EFEFEF} \multicolumn{6}{|c|}{Number of Shots = 1} \\
\hhline{======}
  Meta-SGD~\citep{li2017meta} & 55.58 $\pm$ 1.43   & 32.38 $\pm$ 1.32   & 52.99 $\pm$ 1.36   & 41.74 $\pm$ 1.34   & 45.67\\
  
  \hhline{=::=::=::=::=::=:}
  MAML~\citep{finn2017model} &
  53.94 $\pm$ 1.45   & 31.66 $\pm$ 1.31   & 51.37 $\pm$ 1.38   & 42.12 $\pm$ 1.36   & 44.77\\
  
  \hhline{=::=::=::=::=::=:}
  MT-Net~\citep{lee2018gradient} &
  58.72 $\pm$ 1.43   & 32.80 $\pm$ 1.35   & 47.72 $\pm$ 1.46   & 43.11 $\pm$ 1.42   & 45.59\\
  
  \hhline{=::=::=::=::=::=:}
  B-MAML~\citep{yoon2018bayesian} &
   54.89 $\pm$ 1.48   & 32.53 $\pm$ 1.33   & 53.63 $\pm$ 1.37   & 42.50 $\pm$ 1.33   & 45.88\\
  
  \hhline{=::=::=::=::=::=:}
  HSML~\citep{yao2019hierarchically} &
  55.99 $\pm$ 1.41   & 32.51 $\pm$ 1.35   & 51.26 $\pm$ 1.35   & 42.86 $\pm$ 1.42   & 45.66\\
  
    \hhline{=::=::=::=::=::=:}
  MuMo-MAML~\citep{vuorio2019multimodal} &
  56.82 $\pm$ 1.49   & 33.81 $\pm$ 1.36   & 53.14 $\pm$ 1.39   & 42.22 $\pm$ 1.40   & 46.50 \\
  
  \hhline{=::=::=::=::=::=:}
  ARML~\citep{yao2020automated} &
  59.43 $\pm$ 1.46   & 33.30 $\pm$ 1.30   & 56.20 $\pm$ 1.34   & 45.85  $\pm$ 1.46 & \cellcolor[HTML]{FFDD86}48.70
  \\
  
  \hhline{=::=::=::=::=::=:}
  \textbf{Proposed} &
  59.71 $\pm$ 1.46 &
  35.47 $\pm$ 1.38 &
  57.55 $\pm$ 1.37&
  44.97 $\pm$ 1.44&
  \cellcolor[HTML]{b1e9b0}49.425\\
  
  \hhline{======}
\rowcolor[HTML]{EFEFEF} \multicolumn{6}{|c|}{Number of Shots = 5} \\
\hhline{======}

Meta-SGD~\citep{li2017meta} & 67.87 $\pm$ 0.74	  & 45.49 $\pm$ 0.68   & 66.84 $\pm$ 0.70	& 52.51 $\pm$ 0.81	 & 58.18 \\
  
  \hhline{=::=::=::=::=::=:}
  MAML~\citep{finn2017model} &
  68.52 $\pm$ 0.79	  & 44.56 $\pm$ 0.68   & 66.18 $\pm$ 0.71	& 51.85 $\pm$ 0.85	 & 57.77\\
  
  \hhline{=::=::=::=::=::=:}
  MT-Net~\citep{lee2018gradient} &
  69.22 $\pm$ 0.75	  & 46.57 $\pm$ 0.70   & 63.03 $\pm$ 0.69	& 53.49 $\pm$ 0.83	 & 58.08\\
  
  \hhline{=::=::=::=::=::=:}
  B-MAML~\citep{yoon2018bayesian} &
   69.01 $\pm$ 0.74	  & 46.06 $\pm$ 0.69   & 65.74 $\pm$ 0.67	& 52.43 $\pm$ 0.84	 & 58.31 \\
  
  \hhline{=::=::=::=::=::=:}
  HSML~\citep{yao2019hierarchically} &
  72.07 $\pm$ 0.71	  & 44.71 $\pm$ 0.66   & 64.73 $\pm$ 0.69	& 53.38 $\pm$ 0.79   & 58.65\\
  
    \hhline{=::=::=::=::=::=:}
  MuMo-MAML~\citep{vuorio2019multimodal} &
  70.49 $\pm$ 0.76	  & 45.89 $\pm$ 0.69   & 67.31 $\pm$ 0.68	& 53.96 $\pm$ 0.82	 & 59.41  \\

  \hhline{=::=::=::=::=::=:}
  ARML~\citep{yao2020automated} &
  71.97 $\pm$ 0.70	  & 47.18 $\pm$ 0.78   & 73.63 $\pm$ 0.64	& 55.23 $\pm$ 0.81  & \cellcolor[HTML]{FFDD86}62.00 \\
   
   \hhline{=::=::=::=::=::=:}
  \textbf{Proposed} &
  73.09 $\pm$ 0.73 &
  48.62 $\pm$ 0.69 &
  72.88 $\pm$ 0.64 &
  56.11 $\pm$ 0.81 &
  \cellcolor[HTML]{b1e9b0}62.675\\
  
  \hhline{|-||-||-||-||-||-|}

\end{tabular}
}
\label{table:task-shift}
\end{table*}

\begin{table*}[t]
\centering
\caption{\textbf{Multi-Domain task adaptation}. Performance comparison of ARML and \name when the meta-learners were trained using tasks from multiple domains. \name produces consistently improved generalization in all settings.}
\renewcommand*{\arraystretch}{1.3}
\resizebox{0.85\textwidth}{!}{
\begin{tabular}{|c||c||c||c||c||c|}
  \hhline{|-||-||-||-||-||-|}
  \cellcolor[HTML]{C0C0C0}Method& 
  \cellcolor[HTML]{C0C0C0}ClipArt&
  \cellcolor[HTML]{C0C0C0}InfoGraph&
  \cellcolor[HTML]{C0C0C0}Painting & 
  \cellcolor[HTML]{C0C0C0}QuickDraw & 
  \cellcolor[HTML]{C0C0C0}Average\\
    
\hhline{======}
\rowcolor[HTML]{EFEFEF} \multicolumn{6}{|c|}{Number of Shots = 1} \\
\hhline{======}

ARML~\citep{yao2020automated} &
  47.46 $\pm$ 1.48	  & 30.61 $\pm$ 1.26 & 40.26 $\pm$ 1.41	& 65.71 $\pm$ 1.33	 & \cellcolor[HTML]{FFDD86}46.01 
  \\
  
  \hhline{=::=::=::=::=::=:}
  \textbf{Proposed} &
  49.52 $\pm$ 1.47	  & 34.96 $\pm$ 1.39 & 43.98 $\pm$ 1.47	& 63.59 $\pm$ 1.40 &
  \cellcolor[HTML]{b1e9b0}48.01\\ 
   \hhline{======}

\rowcolor[HTML]{EFEFEF}\multicolumn{6}{|c|}{Number of Shots = 5} \\

\hhline{======}
ARML~\citep{yao2020automated} &
  66.58 $\pm$ 0.73	  & 46.19 $\pm$ 0.76  & 56.86 $\pm$ 0.72	& 83.14 $\pm$ 0.55	 & \cellcolor[HTML]{FFDD86}63.19
  \\
  \hhline{=::=::=::=::=::=:}
  \textbf{Proposed} &
  68.47 $\pm$ 0.71	  & 50.35 $\pm$ 0.75 & 60.94 $\pm$ 0.70	& 83.47 $\pm$ 0.57 &
  \cellcolor[HTML]{b1e9b0}65.80\\
  
  \hhline{|-||-||-||-||-||-|} 
\end{tabular}
}
\label{table:multi-domain}
\end{table*}

\paragraph{\textbf{Datasets}.} We consider two large-scale benchmark datasets to evaluate our proposed task-aware modulation approach under the three settings in Figure \ref{fig:problems}: \textit{(i) Meta-Dataset:} This is a widely adopted benchmark~\citep{triantafillou2019meta} for few-shot image classification and is comprised of multiple image-classification datasets. From this benchmark, we utilize six datasets for our experiments - (a) CUB-200-2011 (Bird) dataset with 200 classes; (b) describable textures dataset (Texture) with 43 classes; (c) FGVC aircraft (Aircraft) dataset with 100 classes; (d) FGVCx-fungi (Fungi) dataset with 1500 classes; (e) VGG flowers (Flower) dataset containing 102 classes; and (f) German traffic signs dataset (Traffic) with 43 classes. For training \name~as well as the baseline methods, we sampled $5$-way few-shot tasks from these datasets for $1-$ and $5-$shot training settings. In each of these datasets, we also constructed disjoint subsets of classes $\mathcal{C}^{tr}$ and $\mathcal{C}^{te}$ for training and testing respectively. For evaluation, we again construct $k$-shot $N$-way tasks from the unseen classes $\mathcal{C}^{te}$. For training and evaluation following SOTA, the images were sized $84\times84\times3$; \textit{(ii) DomainNet:} This is a popular benchmark dataset~\citep{peng2019moment} for domain adaptation and contains images from six different domains (clip-art, info-graph, painting, quick-draw, real, and sketch) belonging to $345$ classes. To ensure availability of sufficient data for creating few-shot tasks, we ignored classes which have less then 50 images and used random splits of $136$ and $39$ classes for training and evaluation.

\noindent \textbf{Experimental details:} For all our experiments we utilized a meta knowledge graph with 4 nodes each of size 128. We leverage a single layer Graph Convolutional Network (GCN) with \textit{tanh} activation for neural message passing. The base learner used a 4 layer CNN with $3\times3$ filters and a single liner classification layer. The 1-shot algorithms were trained for 50K iterations and the 5-shot experiments we trained for 40K iterations, both using a meta batch size of $4$. We utilized the Adam optimizer for the meta update step and for the inner loop, we performed 5 gradient steps using SGD.


\subsection{Findings}
\noindent \textbf{\name~is an effective protocol for task-aware modulation.} In our first experiment, we evaluated the ability of \name~to adapt to novel tasks sampled from unseen classes and compared against different gradient-based meta learning approaches on the Meta-Dataset benchmark. From the results in Table \ref{table:task-shift}, we clearly notice that approaches that leverage task-aware modulation, e.g., MuMo-MAML, HSML, ARML, \name~etc., consistently outperform vanilla meta-learning approaches such as MAML and Meta-SGD. Among existing task-aware modulation strategies, ARML has been known to produce state-of-the-art results on this benchmark\footnote{We used the official implementation from the original authors (https://github.com/huaxiuyao/ARML) to generate all results in our experimental setup. Even with the prescribed settings, we are not able to reproduce the results from their paper. A few others have also raised this issue on Github, but the authors had not responded at the time of submission.}. We find that \name~provides marginal improvements over ARML in both $1-$ and $5-$ shot training settings as shown in Table \ref{table:task-shift}. This can be attributed to the ability of \name~to capture complex task relations using the meta knowledge structure and to effectively distill relevant historical information into the embedding function. Interestingly, CAML eliminates the need for sophisticated task encoding architectures typically used by existing approaches (e.g., RNN autoencoders).

\begin{table*}[t]
\centering
\caption{\textbf{Dataset Generalization}. The evaluation is carried out using a leave-one-out protocol on the meta-dataset. We find that, \name~achieves significantly improved performance over ARML.}
\renewcommand*{\arraystretch}{1.3}
\resizebox{0.95\textwidth}{!}{
\begin{tabular}{|c||c||c||c||c||c||c||c|}
  \hhline{|-||-||-||-||-||-||-||-|}
  \cellcolor[HTML]{C0C0C0}Method& 
  \cellcolor[HTML]{C0C0C0}Bird&
  \cellcolor[HTML]{C0C0C0}Texture&
  \cellcolor[HTML]{C0C0C0}Aircraft & 
  \cellcolor[HTML]{C0C0C0}Fungi & 
  \cellcolor[HTML]{C0C0C0}Flower&
  \cellcolor[HTML]{C0C0C0}Traffic&
  \cellcolor[HTML]{C0C0C0}Average\\

  \hhline{========}
\rowcolor[HTML]{EFEFEF} \multicolumn{8}{|c|}{Number of Shots = 1} \\
\hhline{========}
  ARML~\citep{yao2020automated} &
  38.34 $\pm$ 1.35 & 
  27.13 $\pm$ 1.33 & 
  27.45 $\pm$ 1.23 & 
  32.85	$\pm$ 1.38 & 
  54.79 $\pm$ 1.35 & 
  39.36 $\pm$ 1.33 & \cellcolor[HTML]{FFDD86}36.65
  \\
  
  \hhline{=::=::=::=::=::=::=::=:}
  \textbf{Proposed} &
   40.56 $\pm$ 1.42 & 
   28.75 $\pm$ 1.33 & 
   28.41 $\pm$ 1.24 & 
   33.73 $\pm$ 1.37 & 
   57.89 $\pm$ 1.43 & 
   44.22 $\pm$ 1.39 & \cellcolor[HTML]{b1e9b0}38.93\\
   \hhline{========}

\rowcolor[HTML]{EFEFEF}\multicolumn{8}{|c|}{Number of Shots = 5} \\

\hhline{========}
ARML~\citep{yao2020automated} &
  55.48	$\pm$ 0.80&
  36.49 $\pm$ 0.64&
  36.39 $\pm$ 0.63&
  44.15	$\pm$ 0.73&
  71.80 $\pm$ 0.68&
  52.69 $\pm$ 0.66&\cellcolor[HTML]{FFDD86}49.5
  \\
  
  \hhline{=::=::=::=::=::=::=::=:}
  \textbf{Proposed} &
  58.48 $\pm$ 0.72& 
  39.78 $\pm$ 0.65& 
  39.45 $\pm$ 0.65& 
  45.26 $\pm$ 0.75& 
  73.12 $\pm$ 0.69& 
  62.18 $\pm$ 0.69& \cellcolor[HTML]{b1e9b0}53.1
  \\ 
  \hhline{|-||-||-||-||-||-||-||-|}
\end{tabular}
}
\label{table:dataset-generalization}
\end{table*}



\noindent \textbf{\name~is consistently better in multi-domain task adaptation.} To further study the performance of \name~on heterogeneous task distributions, in this experiment, we considered DomainNet, a multi-domain benchmark. While both the training and testing tasks were drawn from the same collection of domains (ClipArt, InfoGraph, Painting, QuickDraw), we ensured that the set of classes $\mathcal{C}^{tr}$ and $\mathcal{C}^{te}$ were disjoint. In this setting, the increased complexity of the task distribution makes the modulation process more sensitive, when compared to the previous experiment. For simplicity, we compare \name~with the best performing task-aware modulation baseline, \textit{i.e.}, ARML. As shown in Table \ref{table:multi-domain}, CAML achieves performance gaps of $2\%$ and $2.6\%$ on average, in $1-$ and $5-$shot settings respectively. This clearly demonstrates the effectiveness of \name~in handling heterogeneous task distributions.

\begin{figure*}[t]
    \centering
    \includegraphics[width=0.85\textwidth]{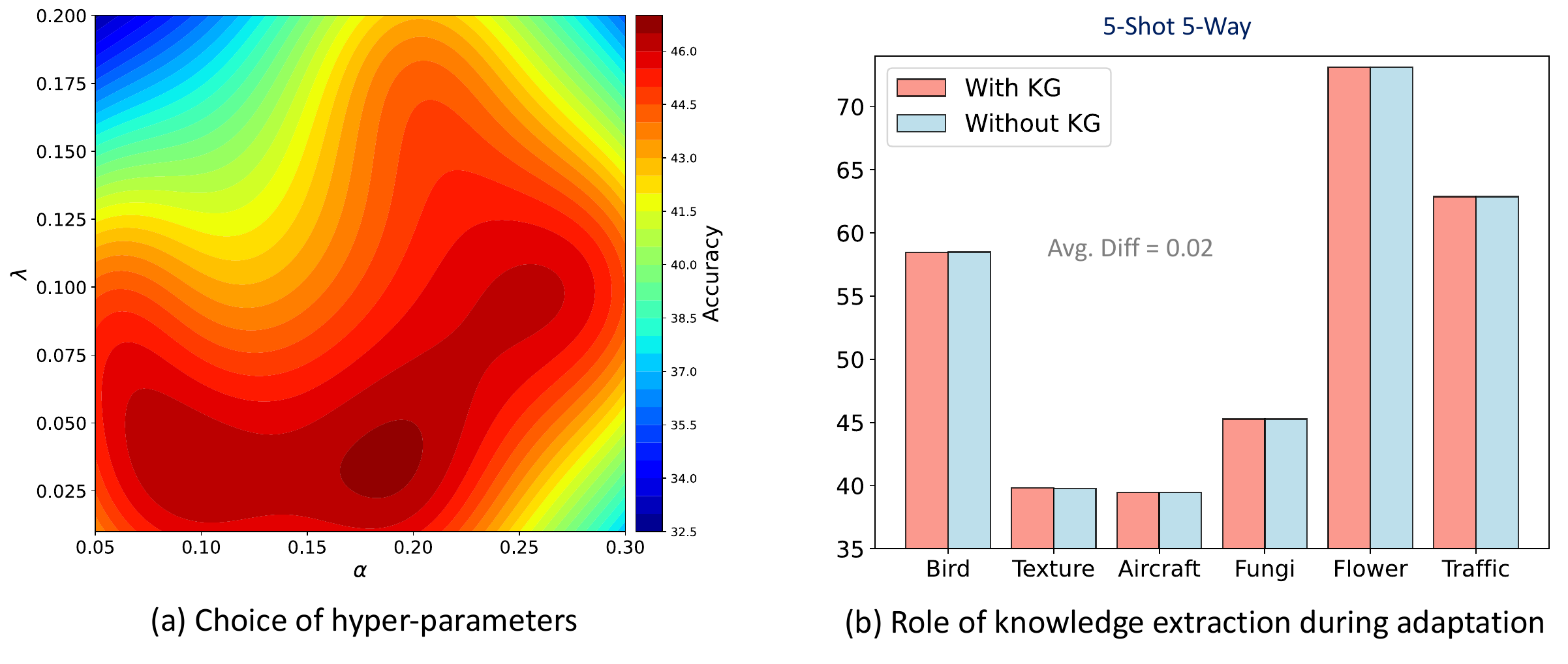}
    \caption{\textbf{Ablations}. (a) Sensitivity of $\alpha$, $\lambda$; (b) We observe that the adaptation performance of \name~does not improve by including the meta knowledge graph (KG) during adaptation.}
    \label{fig:comp}
\end{figure*}

\noindent \textbf{\name~provides significant gains in dataset generalization.}
Finally, the dataset generalization experiment investigates the ability of \name to generalize to unseen datasets - for example utilizing the knowledge about \textit{Fungi} and \textit{Birds} to solve tasks sampled from \textit{Traffic}. The lack of apparent semantic similarity between the classes across different datasets can make this significantly harder than the previous two scenarios. However, improved performance in this problem will be of the most practical value.

In this experiment, we evaluate the generalization using a leave-one-out protocol, where we train the meta learner using $5$ datasets in Meta-Dataset and evaluate on the sixth dataset. From Table \ref{table:dataset-generalization}, we find that \name~achieves significant performance gains over ARML for all datasets and training settings. For example, on the \textit{Traffic} dataset, the performance gain is as high as $\sim5\%$ in the $1-$shot case and $\sim10\%$ in the $5-$shot case, thus establishing \name~as the best performing task-aware modulation approach. The observed performance improvements emphasize the efficacy of our meta knowledge construction process, and the expressiveness of the task representations learned by \name.


\subsection{Ablations}
In this section, we briefly discuss about the behavior of \name~and the impact of different design choices. 

\noindent (i) \textbf{\textit{Choice of $\alpha$ and $\lambda$}}: Figure \ref{fig:comp}(a) illustrates the sensitivity of different choices for $\alpha$ and $\lambda$. While $\alpha$ controls the degree to which the history is retained, $\lambda$ controls the penalty for the distillation cost. These two parameters are used to trade-off generalization (to new tasks) and customization (to observed tasks) of the learner. We find that, when $\alpha$ is very low, i.e., knowledge graphs evolves slowly, using a higher $\lambda$ hurts the performance. On the other hand, for a reasonably higher $\alpha = 0.2$, the choice of $\lambda$ becomes less sensitive. In all our experiments, we used $\alpha = 0.2, \lambda=0.05$.

\noindent (ii) \textbf{\textit{Choice of feature extractor}}: CAML uses an image embedding function to create node features for the prototype graph, which is then subsequently used to generate a task representation. In the algorithm described in the main paper, this embedding module is implemented using a convolutional neural network. As discussed in the ablations (Section 5.2), we studied the impact of the choice of architecture for this module. In particular, we experimented with (i) ResNet-18 model; and (ii) a shallow CNN model (similar to MuMo-MAML), for the case of dataset generalization. As illustrated in Table \ref{table:ablation-feature_extractor}, we find that the performance gap between the two models is only $\sim 0.8\%$ on average, in terms of generalizing to unseen datasets. This behavior emphasizes the flexibility of implementing CAML in practice, wherein the proposed contrastive distillation strategy is effective with any architecture choice.

\begin{table}[t]
\centering
\caption{\textbf{Choice of architecture for the image embedding module}. We compare the performance of a shallow CNN model and ResNet-18 (used in the main paper) using the dataset generalization experiment. In each case, we show the results obtained using $1000$ test tasks for the $1-$shot setting. }
\renewcommand*{\arraystretch}{1.3}
\resizebox{0.5\columnwidth}{!}{
\begin{tabular}{|c||c||c|}
  \hhline{|-||-||-|}
  \cellcolor[HTML]{C0C0C0}Dataset& 
  \cellcolor[HTML]{C0C0C0}ResNet-18&
  \cellcolor[HTML]{C0C0C0}Shallow CNN\\

  \hhline{===}
  Bird & 40.56 $\pm$ 1.42 & 38.18 $\pm$ 1.34\\
  Texture & 28.75 $\pm$ 1.33 &27.91 $\pm$ 1.33 \\
  Aircraft & 28.41 $\pm$ 1.24 &27.77 $\pm$ 1.27 \\
  Traffic & 44.22 $\pm$ 1.39 & 44.70 $\pm$ 1.32 \\
  \hhline{===}
  Average & 35.48 & 34.64\\
  \hhline{|-||-||-|}
\end{tabular}
}
\label{table:ablation-feature_extractor}
\end{table}

\begin{table}[t]
\centering
\caption{\textbf{Choice of task encoding scheme}. We compare the performance of a sophisticated task encoding process (RNN autoencoder) and the na\"ive average pooling strategy used in the main paper, using the dataset generalization experiment. In each case, we show the results obtained using $1000$ test tasks for the $1-$shot setting. }
\renewcommand*{\arraystretch}{1.3}
\resizebox{0.5\columnwidth}{!}{
\begin{tabular}{|c||c||c|}
  \hhline{|-||-||-|}
  \cellcolor[HTML]{C0C0C0}Dataset& 
  \cellcolor[HTML]{C0C0C0}Average Pooling&
  \cellcolor[HTML]{C0C0C0}RNN Autoenc.\\

  \hhline{===}
  Bird & 40.56 $\pm$ 1.42 &40.88 $\pm$ 1.39\\
  Texture & 28.75 $\pm$ 1.33 &28.41 $\pm$ 1.28 \\
  Aircraft & 28.41 $\pm$ 1.24 &28.75 $\pm$ 1.25 \\
  Fungi & 33.73 $\pm$ 1.37 & 34.71	$\pm$ 1.38\\
  Flower & 57.89 $\pm$ 1.43 & 57.53 $\pm$  1.44\\
  Traffic & 44.22 $\pm$ 1.39 &43.44 $\pm$ 1.35\\
  \hhline{===}
  Average & 38.93 & 38.95 \\
  \hhline{|-||-||-|}
\end{tabular}
}
\label{table:ablation-feature_extractor}
\end{table}

\noindent (iii) \textbf{\textit{Choice of task encoding}}: The task encoder in CAML is used generate a compact representation for a given task, based on its prototype graph features. Though a contrastive training objective is included during training, the prototype node features from the image embedding module can be directly used for constructing task representations. In existing task-aware MAML approaches, this step has been found to be crucial and hence sophisticated task encoders are often used. For example, state-of-the-art approaches such as HSML and ARML utilize a RNN auto-encoder, wherein the latent representations from the bottleneck layer are used as the task encodings. In our paper, we argued that, through the use of inherently effective image embeddings (a result of our proposed contrastive training process), we are able to use na\"ive task encodings without any learnable parameters. To validate this claim, we re-implemented CAML using the RNN autoencoder-based task encodings and compared it against the results reported in the main paper. Similar to the previous ablation, we used the dataset generalization experiment and report results for the $1-$shot setting (see Table \ref{table:ablation-feature_extractor}). We find that the RNN autoencoder did not lead to any significant changes in performance (on average the difference was only $0.02$\%).

\noindent (iv) \textbf{\textit{Influence of using meta knowledge during adaptation}}: Though we used a simple protocol for adaptation, we also experimented a variant, where we performed knowledge infusion (using NMP) at test time. As showed in Figure \ref{fig:comp}(b), we found that this did not provide any additional performance gains, thus implying that the relevant prior information has already been effectively distilled into the embedding function. 

\begin{figure}[t]
    \centering
    \includegraphics[width=0.6\columnwidth]{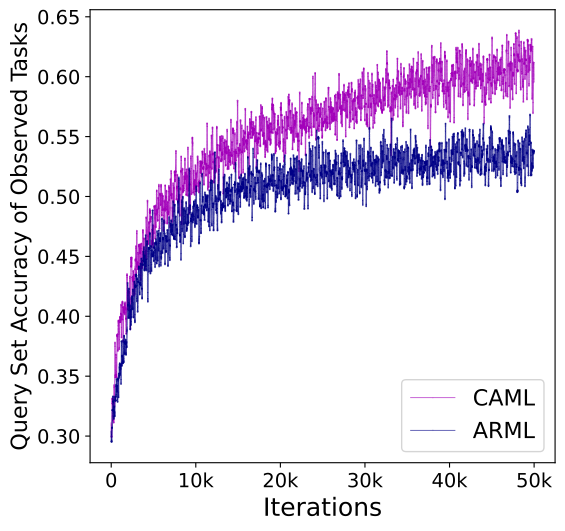}
    \caption{Convergence characteristics of CAML and ARML for a dataset generalization experiment in the $1-$shot training setting.}
    \label{fig:conv}
\end{figure}

\subsection{Analysis}
\noindent \textbf{Training Behavior.} We now report the training behavior of CAML, in comparison to ARML, in terms of the accuracy metric measured using the query sets of each of the training tasks observed during every iteration of training. Figure \ref{fig:conv} shows the convergence characteristics of the two methods for the dataset generalization case, where we train the models using the following $5$ datasets. Bird, Texture, Aircraft, Fungi, Flower. As expected, the inherently large diversity in the observed tasks from the different datasets makes training of both CAML and ARML challenging. Note, these results correspond to the $1-$shot training case. From the plot, it is apparent that, CAML demonstrates improved convergence and produces higher accuracies on the query set, even in this challenging setting.

\begin{figure}[t]
    \centering
    \includegraphics[width=0.75\columnwidth]{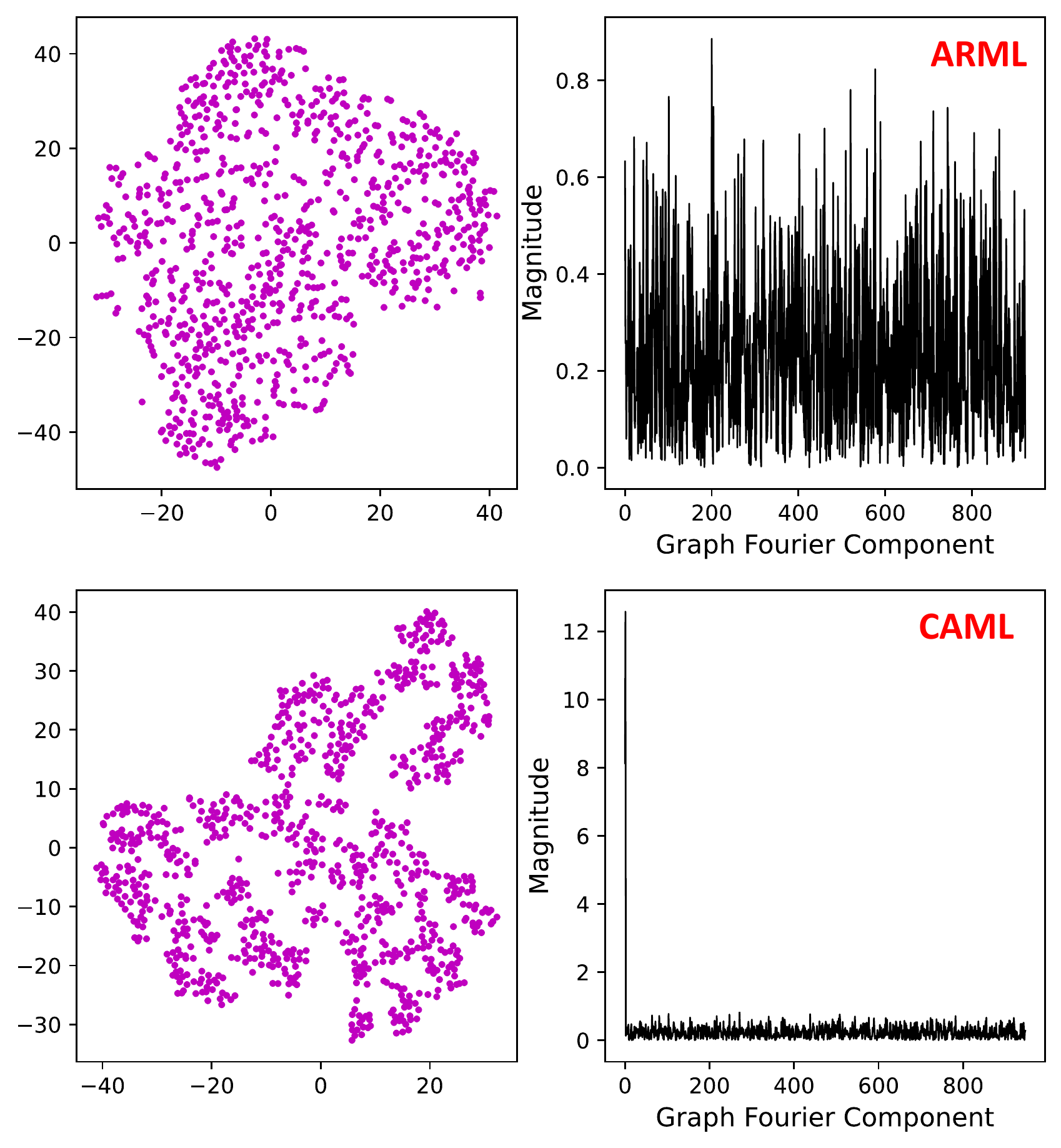}
    \caption{Graph Signal Analysis of the task encodings from CAML (bottom) and ARML (top). For each method, we show the $2-$D TSNE embeddings of the corresponding task encodings and the graph Fourier spectra of the accuracy scores on the k-nearest neighbor graphs constructed from the task encodings ($k=5$).}
    \label{fig:gsp}
\end{figure}

\noindent \textbf{Analysis of task encodings.} In order to analyze the characteristics of the task encodings from CAML and ARML, we conducted the following analysis using tools from graph signal processing. More specifically, we first computed the set of task representations $\mathrm{Z}^{\text{CAML}}$ and $\mathrm{Z}^{\text{ARML}}$ respectively, for a set of $1000$ unseen tasks from a dataset generalization experiment (Traffic is the unseen dataset). We also obtain the few-shot classification accuracies for all $1000$ tasks using the CAML and ARML models (on the query set), which are denoted as $\mathrm{f}^{\text{CAML}}$ and $\mathrm{f}^{\text{ARML}}$. 

Our hypothesis is that if the task representations are informative, two different tasks that have similar encodings (and hence similar task characteristics) should lead to similar accuracy scores. To test this hypothesis, we construct k-nearest neighbor graphs for both CAML and ARML embeddings to obtain the graph adjacency matrices $\mathrm{G}^{\text{CAML}}$ and $\mathrm{G}^{\text{ARML}}$. Next, we computed the graph Fourier basis using the pygsp package (https://github.com/epfl-lts2/pygsp). Finally, we compute the graph Fourier transform for the function defined as the accuracy scores. The expectation is that, when the task encodings are meaningful, the resulting graph Fourier spectrum should concentrate most of the signal's energy at low-frequencies. Figure \ref{fig:gsp} illustrates the Fourier spectra obtained for results from CAML and ARML, when the number of neighbors $k$ was set to $5$. Even with such a small neighborhood size, the spectra for ARML contains non-trivial energy at even high frequencies, thus indicating that the task encodings are not  consistent with the expected classification performance. In contrast, for CAML, we notice that most of the signal energy is concentrated at low frequencies, thus justifying its improved generalization.

\section{Conclusions}
In this work, we presented \name, a novel knowledge-enhanced meta-learning approach for few-shot classification tasks. \name~employs a knowledge extraction process that distills prior task information from the learnable knowledge structure to the embedding function using a contrastive objective. This eliminates the need for using the knowledge structure during adaptation and is able to aptly modulate the meta-initialization parameters solely using the task representations obtained by a simple average pooling of the prototype embeddings. Using extensive empirical studies on different task-adaptation settings, we showed that \name~consistently outperforms existing baselines. This work clearly establishes \name~as the best performing task-aware modulation technique and further motivates the study of constructing generalizable knowledge priors for few-shot task adaptation under challenging distribution shifts.

\bibliography{main}


\end{document}